\documentclass{article}

\newlength{\lengthcorrection}

\def\prod{\ensuremath{\Rightarrow}}
\def\tuple#1{\ensuremath{\langle#1\rangle}}

\def\|{\ensuremath{\mid}}

\def\skipone#1{} 
\def\realstring#1{
        \expandafter\skipone\string#1}
\def\ifundefined#1{%
        \expandafter\ifx\csname\realstring{#1}\endcsname\relax}

\def\os#1{``{\em #1}\/''}

\def\svar{\rule{1.8ex}{0.2ex}}
\def\mvar{\ensuremath{\mathbf =}}

\def\C{\ensuremath{\mathcal C}}

\def\pre{\ensuremath{\prec}}
\def\preD{\ensuremath{\ll}}

\newcounter{example} 
\renewcommand{\theexample}{(\arabic{example})}
\newcounter{subexample}[example]

\def\ex#1{
	\par\addvspace{1em}
	\begin{list}{}{	\setlength{\itemsep}{0pt}
			\setlength{\parsep}{0pt}
			}
		\refstepcounter{example}
		\setlength{\topsep}{0pt} 
		\item[\theexample] #1 
	\end{list}
	\par\addvspace{1em}}

\def\*{\raisebox{-0.3ex}{*}}

\def\.{$\circ$}

\let\orig=\|

\usepackage{aclap-c}

\usepackage{array}

\usepackage{latexsym}
\usepackage{epsfig}

\usepackage[leqno]{amsmath}

\renewcommand{\|}{\orig}

\usepackage{theorem}

\theorembodyfont{\rmfamily}
\theoremstyle{break}
\newtheorem{definition}{Definition}
\newtheorem{lemma}{Lemma}
\newtheorem{theorem}[lemma]{Theorem}

\usepackage{nobibib-f}

\renewcommand{\v}[1]{$\check{\text{#1}}$}

\newcommand{\meth}[1]{\ensuremath{\text{{\small \sffamily #1}}}}

\newcommand{\intr}[1]{\textsl{#1}}

\newcommand{\defed}{\ensuremath{\hfill \lhd}}
\newcommand{\proofed}{\hfill \rule{2mm}{2mm}}
\newcommand{\said}{\ensuremath{\hfill \Box}}

\newcommand{\mycirc}{\ensuremath{\mbox{{\rm \lower-.4ex\hbox{{\scriptsize $\bigcirc$}}}}}}

\newcommand{\mala}[1]{}

\def\thisversion{$ $Revision: 2.3 $ $}
\def\thisdate{$ $Date: 1997/04/07 16:56:39 $ $}

\begin{document}

\title{\vspace{-0.5in}\bf 
	The Complexity of Recognition \\ 
	of Linguistically Adequate Dependency Grammars
	\thanks{\ Appeared In: Proc. 35th Annual Meeting of the ACL and 8th
	Conf. of the EACL, Madrid 1997, pp.337--343.}
	}

\author{\bf{Peter Neuhaus} \\
	\bf{Norbert Br\"{o}ker} \\
	{\small \raisebox{-0.25mm}{\epsfxsize=5ex\epsfbox{clif-logo-c.eps}}}
		Computational Linguistics Research Group\\
	Freiburg University, Friedrichstra\ss e 50\\
	D-79098 Freiburg, Germany\\
	\small{email: \{neuhaus,nobi\}@coling.uni-freiburg.de}\\
}


\date{}


\maketitle
\vspace{-0.5in}
\begin{abstract}
Results of computational complexity exist for a wide range of phrase
structure-based grammar formalisms, while there is an apparent lack of such
results for dependency-based formalisms. We here adapt a result on the
complexity of ID/LP-grammars to the dependency framework. Contrary to
previous studies on heavily restricted dependency grammars, we prove
that recognition (and thus, parsing) of linguistically adequate dependency 
grammars is $\mathcal{NP}$-complete.
\end{abstract}

\pagestyle{empty}

%
%

\def\thisversion{$ $Revision: 1.13 $ $}
\def\thisdate{$ $Date: 1996/12/31 11:38:08 $ $}


\section{Introduction
	\label{ch:intro}}

The introduction of dependency grammar (DG) into modern linguistics is
marked by \textcite{Tesniere1959}. His conception addressed didactic goals
and, thus, did not aim at formal precision, but rather at an intuitive
understanding of semantically motivated dependency relations. An early
formalization was given by \textcite{gaifman65}, who showed the generative
capacity of DG to be (weakly) equivalent to standard context-free
grammars. Given this equivalence, interest in DG as a linguistic framework
diminished considerably, although many dependency grammarians view
Gaifman's conception as an unfortunate one (cf. Section \ref{ch:gaifman}).
To our knowledge, there has been no other formal study of DG.%
This is reflected by a recent study \cite{lombardo+lesmo96}, which applies
the Earley parsing technique \cite{earley70} to DG, and thereby
achieves cubic time complexity for the analysis of DG. In their discussion,
\citeauthor{lombardo+lesmo96} express their hope that slight
increases in generative capacity will correspond to equally slight
increases in computational complexity. It is this claim that we challenge
here.

After motivating non-projective analyses for DG, we investigate various
variants of DG and identify the separation of dominance and precedence as a
major part of current DG theorizing. Thus, no current variant of DG (not
even \citeauthor{Tesniere1959}'s original formulation) is compatible with
Gaifman's conception, which seems to be motivated by formal considerations
only (viz.,\ the proof of equivalence). Section \ref{ch:grammar} advances
our proposal, which cleanly separates dominance and precedence
relations. This is illustrated in the fourth section, where we give a
simple encoding of an $\mathcal{NP}$-complete problem in a discontinuous
DG. Our proof of $\mathcal{NP}$-completeness, however, does not rely on
discontinuity, but only requires unordered trees. It is adapted from a
similar proof for unordered context-free grammars (UCFGs) by
\textcite{barton85}.

%
%

\def\thisversion{$ $Revision: 1.17 $ $}
\def\thisdate{$ $Date: 1997-04-07 18:34:33+02 $ $}


\section{Versions of Dependency Grammar
	\label{ch:DG}}

The growing interest in the dependency concept (which roughly corresponds
to the $\theta$-roles of GB, subcategorization in HPSG, and the so-called
domain of locality of TAG) again raises the issue whether non-lexical
categories are necessary for linguistic analysis. After reviewing several
proposals in this section, we argue in the next section that word order ---
the description of which is the most prominent difference between PSGs and
DGs --- can adequately be described without reference to non-lexical
categories.


Standard PSG trees are {\em projective}, i.e., no branches cross when the
terminal nodes are projected onto the input string. In contrast to PSG
approaches, DG requires non-projective analyses. As DGs are restricted to
lexical nodes, one cannot, e.g., describe the so-called unbounded
dependencies without giving up projectivity. First, the categorial approach
employing partial constituents
\cite{Huck1988,Hepple1990} is not available, since there are no phrasal
categories. Second, the coindexing \cite{Haegeman1994} or structure-sharing
\cite{PS-II} approaches are not available, since there are no empty
categories. 

Consider the extracted NP in \os{Beans, I know John likes} (cf. also to
Fig.\ref{fig:discont} in Section \ref{ch:grammar}). A projective tree would
require \os{Beans} to be connected to either \os{I} or \os{know} -- none of
which is conceptually directly related to \os{Beans}. It is \os{likes} that
determines syntactic features of \os{Beans} and which provides a semantic
role for it. The only connection between \os{know} and \os{Beans} is that
the finite verb allows the extraction of \os{Beans}, thus defining order
restrictions for the NP. This has led some DG variants to adopt a general
graph structure with multiple heads instead of trees. We will refer to DGs
allowing non-projective analyses as {\em discontinuous DGs}.


\textcite{Tesniere1959} devised a bipartite grammar theory which consists
of a dependency component and a translation component ('translation' used
in a technical sense denoting a change of category and grammatical
function). The dependency component defines four main categories and
possible dependencies between them.
What is of interest here is that there is \emph{no mentioning of order} in
\citeauthor{Tesniere1959}'s work. Some practitioneers of DG 
have allowed word order as a marker for translation,
but they do not prohibit non-projective trees.

	\label{ch:gaifman}

\textcite{gaifman65} designed his DG entirely analogous to 
context-free phrase structure grammars. Each word is associated with a
category, which functions like the non-terminals in CFG. He then defines
the following rule format for dependency grammars:

\ex{	$ X (Y_1, \ldots, Y_i, *, Y_{i+1}, \ldots, Y_n) $}

\noindent
This rule states that a word of category $X$ governs words of category
$Y_1, \ldots, Y_n$ which occur in the given order. The head (the word of
category $X$) must occur between the $i$-th and the $(i+1)$-th modifier.
The rule can be viewed as an ordered tree of depth one with node
labels. Trees are combined through the identification of the root of one
tree with a leaf of identical category of another tree.
This formalization is restricted to projective trees with a completely
specified order of sister nodes. As we have argued above, such a
formulation cannot capture semantically motivated dependencies. 




\subsection{Current Dependency Grammars}

Today's DGs differ considerably from Gaifman's conception, and we will very
briefly sketch various order descriptions, showing that DGs generally
dissociate dominance and precedence by some mechanism. All variants share,
however, the rejection of phrasal nodes (although phrasal features are
sometimes allowed) and the introduction of edge labels (to distinguish
different dependency relations).

Meaning-Text Theory
\cite{Melcuk1988} assumes seven strata of
representation. The rules mapping from the unordered dependency trees of
surface-syntactic representations onto the annotated lexeme sequences of
deep-morphological representations include global ordering rules which allow
discontinuities. These rules have not yet been formally specified
\cite[p.187f]{Melcuk+Pertsov1987}, but see the proposal by
\textcite{Rambow+Joshi1994}.

Word Grammar
\cite{Hudson1990} is based on general graphs.  The ordering
of two linked words is specified together with their dependency relation,
as in the proposition ``\meth{object of verb succeeds it}''.  Extraction is
analyzed by establishing another dependency, \meth{visitor}, between the
verb and the extractee, which is required to precede the verb, as in
``\meth{visitor of verb precedes it}''. Resulting inconsistencies, e.g. in
case of an extracted object, are not resolved, however.

Lexicase
\cite{Starosta1988,Starosta1992} employs complex feature structures to
represent lexical and syntactic entities. Its word order description is
much like that of Word Grammar (at least at some level of abstraction), and
shares the above inconsistency.

Dependency Unification Grammar
\cite{Hellwig1988} defines a tree-like data structure for the
representation of syntactic analyses.  Using morphosyntactic features with
special interpretations, a word defines abstract positions into which
modifiers are mapped.  Partial orderings and even discontinuities can thus
be described by allowing a modifier to occupy a position defined by some
transitive head. The approach cannot restrict discontinuities properly,
however.


Slot Grammar
\cite{McCord1990} employs a number of rule types, some of which are
exclusively concerned with precedence. So-called head/slot and slot/slot
ordering rules describe the precedence in projective trees, referring to
arbitrary predicates over head and modifiers. Extractions (i.e.,
discontinuities) are merely handled by a mechanism built into the parser.

This brief overview of current DG flavors shows that various mechanisms
(global rules, general graphs, procedural means) are generally employed to
lift the limitation to projective trees. Our own approach presented below
improves on these proposals because it allows the lexicalized and
declarative formulation of precedence constraints. The necessity of
non-projective analyses in DG results from examples like \os{Beans, I know
John likes} and the restriction to lexical nodes which prohibits
gap-threading and other mechanisms tied to phrasal categories.

\section{A Dependency Grammar with Word Order Domains}
	\label{ch:grammar}

We now sketch a minimal DG that incorporates only word classes and word
order as descriptional dimensions. The separation of dominance and
precedence presented here grew out of our work on German, and retains the
local flavor of dependency specification, while at the same time covering
arbitrary discontinuities. It is based on a (modal) logic with
model-theoretic interpretation, which is presented in more detail in
\cite{nobi-Diss}.

\subsection{Order Specification}

Our initial observation is that DG cannot use binary precedence constraints
as PSG does. Since DG analyses are hierarchically flatter, binary
precedence constraints result in inconsistencies, as the analyses of Word
Grammar and Lexicase illustrate. In PSG, on the other hand, the phrasal
hierarchy separates the scope of precedence restrictions. This effect is
achieved in our approach by defining \emph{word order domains} as sets of
words, where precedence restrictions apply only to words within the same
domain. Each word defines a sequence of order domains, into which the word
and its modifiers are placed.

\begin{figure}
\centering
\epsfig{file=domains.eps}
\caption{Word order domains in \os{Beans, I know John likes}
	\label{fig:discont}}
\end{figure}

Several restrictions are placed on domains. First, the domain sequence must
mirror the precedence of the words included, i.e., words in a prior domain
must precede all words in a subsequent domain.  Second, the order domains
must be hierarchically ordered by set inclusion, i.e., be
projective. Third, a domain (e.g., $d_1$ in Fig.\ref{fig:discont}) can be
constrained to contain at most one partial dependency tree.%
\footnote{
For details, cf. \cite{nobi-Diss}. 
}
We will write singleton domains as ``\svar'', while other domains are
represented by ``\mvar''.  The precedence of words within domains is
described by binary precedence restrictions, which must be locally
satisfied in the domain with which they are associated. Considering
Fig.\ref{fig:discont} again, a precedence restriction for \os{likes} to
precede its object has no effect, since the two are in different domains.
The precedence constraints are formulated as a binary relation ``\pre''
over dependency labels, including the special symbol ``\meth{self}''
denoting the head. Discontinuities can easily be characterized, since a
word may be contained in any domain of (nearly) any of its transitive
heads. If a domain of its direct head contains the modifier, a continuous
dependency results. If, however, a modifier is placed in a domain of some
transitive head (as \os{Beans} in Fig.\ref{fig:discont}), discontinuities
occur. Bounding effects on discontinuities are described by specifying that
certain dependencies may not be crossed.%
\footnote{
German data exist that cannot be captured by the (more common) bounding of
discontinuities by nodes of a certain \emph{category}.  
}
For the purpose of this paper, we need not formally introduce the bounding
condition, though. 

A sample domain structure is given in Fig.\ref{fig:discont}, with two
domains $d_1$ and $d_2$ associated with the governing verb \os{know}
(solid) and one with the embedded verb \os{likes} (dashed). $d_1$ may
contain only one partial dependency tree, the extracted phrase. $d_2$
contains the rest of the sentence. Both domains are described by
\ref{ex:domain-spec}, where the domain sequence is represented as
``\preD''. $d_2$ contains two precedence restrictions which require that
\os{know} (represented by \meth{self}) must follow the subject (first
precedence constraint) and precede the object (second precedence
constraint).

\ex{	\label{ex:domain-spec}
	\svar\ \{\} 
	\preD\ 
	\mvar\ \{ (\meth{subject} \pre\ \meth{self}), 
		(\meth{self} \pre\ \meth{object})\}}

\subsection{Formal Description}

The following notation is used in the proof. A lexicon $Lex$ maps words
from an alphabet $\Sigma$ to word classes, which in turn are associated
with valencies and domain sequences. The set $C$ of \emph{word classes} is
hierarchically ordered by a subclass relation
\ex{$ 
\mbox{\meth{isa}}_C \subset C \times C 
$}
A word $w$ of class $c$ inherits the valencies (and domain sequence) from
$c$, which are accessed by 
\ex{$ 
w.\mbox{\meth{valencies}}
$}
A \emph{valency} \tuple{b, d, c} describes a possible dependency relation
by specifying a flag $b$ indicating whether the dependency may be
discontinuous, the dependency name $d$ (a symbol), and the word class $c
\in \C$ of the modifier.  A word $h$ may govern a word $m$ in dependency
$d$ if $h$ defines a valency \tuple{b, d, c} such that $(m\
\mbox{\meth{isa}}_C\ c)$ and $m$ can consistently be inserted into a domain
of $h$ (for $b = -$) or a domain of a transitive head of $h$ (for $b =
+$). This condition is written as 
\ex{$ 
\mbox{\meth{governs}}(h,d,m)
$}
A DG is thus characterized by 
\ex{$ 
G = \tuple{Lex, C, \mbox{\meth{isa}}_C, \Sigma}
$}
The language $L(G)$ includes any sequence of words for which a dependency
tree can be constructed such that for each word $h$ governing a word $m$ in
dependency $d$, \meth{governs}$(h,d,m)$ holds. The modifier of $h$ in
dependency $d$ is accessed by
\ex{$ 
h.\mbox{\meth{mod}}(d)
$}

%
%

\def\thisversion{$ $Revision: 1.14 $ $}
\def\thisdate{$ $Date: 1997-03-24 17:24:30+01 $ $}


\section{The complexity of DG Recognition}

\textcite[p.728]{lombardo+lesmo96} convey their hope that increasing
the flexibility of their conception of DG will ''\ldots imply the
restructuring of some parts of the recognizer, with a plausible
increment of the complexity''. We will show that adding a little
(linguistically required) flexibility might well render recognition
$\mathcal{NP}$-complete. 
To prove this, we will encode the vertex cover problem, which is known to
be $\mathcal{NP}$-complete, in a DG.


\subsection	{Encoding the Vertex Cover Problem in Discontinuous DG}
  \label{ch:moti}

A \intr{vertex cover} of a finite graph is a subset of its vertices
such that (at least) one end point of every edge is a member of that
set. The \intr{vertex cover problem} is to decide whether for a given
graph there exists a vertex cover with at most $k$ elements. The
problem is known to be $\mathcal{NP}$-complete
\cite[pp.53--56]{garey+johnson83}. Fig.~\ref{fig:vertex-graph} gives a
simple example where $\{c,d\}$ is a vertex cover. 

\begin{figure}[hbt]\centering
\epsfig{file=vertex.eps,width=3.8cm}
\caption{Simple graph with vertex cover $\{c, d\}$.}
\label{fig:vertex-graph}
\end{figure}

A straightforward encoding of a solution in the DG formalism
introduced in Section~\ref{ch:grammar} defines a root word $s$ of
class $S$ with $k$ valencies for words of class $O$. $O$ has $|V|$
subclasses denoting the nodes of the graph. An edge is represented by
two linked words (one for each end point) with the governing word
corresponding to the node included in the vertex cover. The
subordinated word is assigned the class $R$, while the governing word
is assigned the subclass of $O$ denoting the node it represents. The
latter word classes define a valency for words of class $R$ (for the
other end point) and a possibly discontinuous valency for another word
of the identical class (representing the end point of another edge
which is included in the vertex cover).  This encoding is summarized
in Table~\ref{tab:vertex-lexikon}.


\begin{table*}
~\hfill
\begin{tabular}{|>{$}l<{$}|>{$}l<{$}|>{$}l<{$}|}\hline
\text{classes} & \text{valencies} & \text{order domain}\\ 
\hline\hline
S & \{\langle -, \meth{ mark1}, O\rangle ,
	\langle -, \meth{ mark2}, O \rangle \}
	& \mvar\{(\meth{self} \prec \meth{mark1}), 
	(\meth{mark1} \prec \meth{mark2}) \}
\\\hline
A \meth{ isa$_C$ } O 
	& \{\langle -, \meth{ unmrk}, R\rangle, 
	\langle +, \meth{ same}, A \rangle \}
	& \mvar\{(\meth{unmrk} \prec  \meth{same}), 
	(\meth{self} \prec \meth{same}) \}
\\\hline
B \meth{ isa$_C$ } O 
	& \{\langle -, \meth{ unmrk}, R\rangle, 
	\langle +, \meth{ same}, B \rangle \}
	& \mvar\{(\meth{unmrk} \prec \meth{same}), 
	(\meth{self} \prec \meth{same}) \}
\\\hline
C \meth{ isa$_C$ } O 
	& \{\langle -, \meth{ unmrk}, R\rangle, 
	\langle +, \meth{ same}, C \rangle  \}
	& \mvar\{(\meth{unmrk} \prec  \meth{same}), 
	(\meth{self} \prec \meth{same}) \}
\\\hline
D \meth{ isa$_C$ } O 
	& \{\langle -, \meth{ unmrk}, R\rangle, 
	\langle +, \meth{ same}, D\rangle \}
	& \mvar\{(\meth{unmrk} \prec  \meth{same}), 
	(\meth{self} \prec \meth{same}) \}
\\\hline
R
	& \{\ \}
	&  \svar\{\ \}
\\\hline
\end{tabular}
\hfill
\begin{tabular}{|>{$}c<{$}|>{$}l<{$}|}
\hline
\text{word} & \text{classes}\\ 
\hline\hline
s & \{S\}\\\hline
a & \{A, R\} \\\hline
b & \{B, R\}\\\hline
c & \{C, R\}\\\hline
d & \{D, R\}\\\hline
\end{tabular}
\hfill~
\caption{Word classes and lexicon for vertex cover problem from 
Fig.~\ref{fig:vertex-graph}}
\label{tab:vertex-lexikon}
\end{table*}

The input string contains an initial $s$ and for each edge the words
representing its end points, e.g. ``saccdadbdcb'' for our example. If
the grammar allows the construction of a complete dependency tree
(cf.\ Fig.~\ref{fig:dgvalenz} for one solution), this encodes a
solution of the vertex cover problem.


\begin{figure}\centering
\epsfig{file=dg-vertex-dis.eps,width=3.97cm}
\caption{Encoding a solution to the vertex cover problem from 
Fig.~\ref{fig:vertex-graph}.}
\label{fig:dgvalenz}
\end{figure}

\subsection	{Formal Proof using Continuous DG}
  \label{ch:proof}

The encoding outlined above uses non-projective trees, i.e., crossing
dependencies. In anticipation of counter arguments such as that the
presented dependency grammar was just too powerful, we will present the
proof using only one feature supplied by most DG formalisms, namely the
free order of modifiers with respect to their head. Thus, modifiers must be
inserted into an order domain of their head (i.e., no $+$ mark in
valencies). This version of the proof uses a slightly more complicated
encoding of the vertex cover problem and resembles the proof by
\textcite{barton85}.

\begin{definition}[Measure]
\label{def:leng}\mala{def:leng}
Let $\|\cdot\|$ be a  measure for the encoded input length
of a computational problem. We require that if $S$ is a set or string
and $k \in \mathbf{N}$ then $|S| \geq k$ implies $\|S\| \geq \|k\|$
and that for any tuple $\| \langle \cdots, x, \cdots
\rangle\| \geq \|x\|$ holds. \defed
\end{definition}

\begin{definition}[Vertex Cover Problem] 
\label{def:vertex-problem} 
A possible instance of the vertex cover problem is a triple  $\langle V,
E, k \rangle$ where $\langle V, E \rangle$ is a finite  graph and $|V| >
k \in \mathbf{N}$. The vertex cover problem is the set $VC$ of all 
instances $\langle V, E, k \rangle$ for which there exists a subset $V'
\subseteq V$ and a function $f: E
\rightarrow V'$ such that $|V'| \leq k$ and $\forall \langle v_m, v_n
\rangle \in E: f(\langle v_m, v_n \rangle) \in \{v_m, v_n\}$.
\defed 
\end{definition}

\begin{definition}[DG recognition problem]
A possible instance of the DG recognition problem is a tuple $\langle
G, \sigma \rangle$ where $G = \langle \mathit{Lex}, C, \meth{ isa$_C$ },
\Sigma\rangle$ is a dependency grammar as defined in Section
\ref{ch:grammar}\mala{ch:grammar} and $\sigma \in \Sigma^+$.  The DG
recognition problem $DGR$ consists of all instances $\langle G, \sigma
\rangle$ such that $\sigma \in L(G)$. \defed
\end{definition}

For an algorithm to decide the $VC$ problem consider a data structure
representing the vertices of the graph (e.g., a set). We separate the
elements of this data structure into the (maximal) vertex cover set
and its complement set. Hence, one end point of every edge is assigned
to the vertex cover (i.e., it is marked). Since (at most) all $|E|$
edges might share a common vertex, the data structure has to be a
\emph{multiset} which contains $|E|$ copies of each vertex.  Thus,
marking the $|V|-k$ complement vertices actually requires marking
$|V|-k$ times $|E|$ identical vertices.  This will leave $(k-1)*|E|$
unmarked vertices in the input structure. To achieve this algorithm
through recognition of a dependency grammar, the marking process will
be encoded as the filling of appropriate valencies of a word $s$ by
words representing the vertices.  Before we prove that this encoding
can be generated in polynomial time we show that:

\begin{lemma}
\label{th:np}\mala{th:np}
The DG recognition problem is in the complexity class $\mathcal{NP}$.
\said
\end{lemma}

Let $G=\langle \mathit{Lex},C, \meth{ isa$_C$ }, \Sigma\rangle$ and
$\sigma \in \Sigma^+$. We give a \emph{nondeterministic} algorithm for
deciding whether $\sigma=\langle s_1 \cdots s_n \rangle$ is in
$L(G)$. Let $H$ be an empty set initially:

\begin{enumerate}
\item Repeat until $|H|=|\sigma|$
\begin{enumerate}
\item 	\begin{enumerate}
	\item For every $s_i \in \sigma$ choose a lexicon entry $c_i \in
	\mathit{Lex}(s_i)$.
	\item From the $c_i$ choose one word as the head $h_0$.
	\item Let $H := \{h_0\}$ and 
	$M := \{c_i | i \in [1,|\sigma|]\}\setminus H$.
	\end{enumerate}
\item Repeat until $M = \emptyset$:
	\begin{enumerate}
	\item Choose a head $h \in H$ and a valency 
	$\langle b, d, c\rangle \in h.\meth{valencies}$ and
	a  modifier $m \in M$.
	\item If $\meth{governs}(h,d,m)$ holds then 
	 establish the dependency relation
	between $h$ and the $m$, and add $m$ to the set $H$.
	\item  Remove $m$ from $M$.
	\end{enumerate}
\end{enumerate}
\end{enumerate}

The algorithm obviously is (nondeterministically) polynomial in the length of
the input. Given that $\langle G, \sigma \rangle \in DGR$, a dependency
tree covering the whole input exists and the algorithm will be able to
guess the dependents of every head correctly. If, conversely, the algorithm
halts for some input $\langle G, \sigma \rangle$, then there necessarily
must be a dependency tree rooted in $h_0$ completely covering
$\sigma$. Thus, $\langle G, \sigma
\rangle \in DGR$. \proofed

\begin{lemma}
\label{th:constr}\mala{th:constr}
Let  $\langle V,E,k \rangle$ be a possible instance of the vertex cover
problem. Then a grammar $G(V,E,k)$ and an input
$\sigma(V,E,k)$ can be constructed in time polynomial in $\|\langle V,E,k
\rangle\|$ such that
\[\langle V,E,k
\rangle \in VC \Longleftrightarrow \langle G(V,E,k),\sigma(V,E,k)
\rangle \in DGR\] 
\said
\end{lemma}

For the proof, we first define the encoding and show that it can be
constructed in polynomial time. Then we proceed showing that the
equivalence claim holds.  The set of classes is $C =_{\text{def}}
\{S,R,U\} \cup \{H_i | i\in[1,|E|]\} \cup \{U_i,V_i |
i\in[1,|V|]\}$. In the \meth{isa}$_C$ hierarchy the classes $U_i$
share the superclass $U$, the classes $V_i$ the superclass $R$.
Valencies are defined for the classes according to
Table~\ref{tab:form}.  Furthermore, we define \(\Sigma =_{\text{def}}
\{s\} \cup \{v_i | i \in [1,|V|]\}\). The lexicon $\mathit{Lex}$
associates words with classes as given in Table~\ref{tab:form}.

\begin{table*}
\begin{tabular}{|>{$}l<{$}|>{$}l<{$}|>{$}l<{$}|>{$}l<{$}|}
\hline
 & \text{word class} & \text{valencies} & \text{order} \\ 
\hline\hline
\forall v_i \in V  & V_i \meth{ isa$_C$ } R & \{\:\} & \mvar\{\:\} \\
\hline
\forall v_i \in V 
	& U_i \meth{ isa$_C$ } U 
	& \{\langle -, \meth{r$_1$}, V_i\rangle, 
	\cdots, 
	\langle -, \meth{r$_{|E|-1}$}, V_i \rangle\}
	& \mvar\{\:\} \\
\hline
\forall e_i \in E 
	& H_i & \{\:\}
	& \mvar\{\:\} \\
\hline
 & S 
	& \{\langle -, \meth{u$_1$}, U\rangle, 
		\cdots, 
		\langle -, \meth{u$_{|V|-k}$}, U\rangle,
 	& \mvar\{\:\} \\
	&	& \langle -, \meth{h$_1$}, H_1\rangle, 
		\cdots, 
		\langle -, \meth{h$_{|E|}$}, H_{|E|}\rangle, & \\
	&	& \langle -, \meth{r$_1$}, R\rangle, 
		\cdots, 
		\langle -, \meth{r$_{(k-1)|E|}$}, R \rangle\} & \\
\hline
\end{tabular}
\hfill
\begin{tabular}{|>{$}c<{$}|l|}
\hline
\text{word} & \text{word classes}\\ 
\hline\hline
v_i & $\{U_i, V_i\} \cup 
	\{H_j |  \exists v_m,v_n \in V:$ \\
    & $\phantom{\{U_i, V_i\} \cup \{H_j |  \exists v} 
	e_j = \langle v_m, v_n\rangle \wedge$\\
    & $\phantom{\{U_i, V_i\} \cup \{H_j |  \exists v} 
	v_i \in \{v_m, v_n\}\}$\\\hline
s & $\{S\}$\\\hline
\end{tabular}
\caption{Word classes and lexicon to encode vertex cover problem}
\label{tab:form}
\end{table*}

We set
\begin{equation*}
G(V,E,k) =_{\text{def}} 
\langle \mathit{Lex}, C, \meth{ isa$_C$ },\Sigma \rangle 
\end{equation*}
and
\begin{equation*}
\sigma(V,E,k) =_{\text{def}} 
s \underbrace{v_1\cdots
v_1}_{|E|}\cdots \underbrace{v_{|V|}\cdots v_{|V|}}_{|E|}
\end{equation*}

For an example, cf.\ Fig.~\ref{fig:disc} which shows a dependency tree
for the instance of the vertex cover problem from
Fig.~\ref{fig:vertex-graph}. The two dependencies $u_1$ and $u_2$
represent the complement of the vertex cover.

\begin{figure}\centering
\epsfig{file=dg-vertex.eps,width=7.5cm}
\caption{Encoding a solution to the vertex cover problem from 
Fig.~\ref{fig:vertex-graph}.}
\label{fig:disc}
\end{figure}

It is easily seen\footnote{The construction requires $2*|V|+|E|+3$
word classes, $|V|+1$ terminals in at most $|E|+2$ readings each. $S$
defines  $|V|+k*|E|-k$ valencies, $U_i$ defines $|E|-1$ valencies. The length
of $\sigma$ is $|V|*|E|+1$.} that
$\|\langle G(V,E,k),\sigma(V,E,k)\rangle\|$ is polynomial in $\|V\|$,
$\|E\|$ and $k$. From $|E| \geq k$ and Definition~\ref{def:leng} it
follows that $\|\langle V,E,k \rangle\| \geq \|E\| \geq \|k\| \geq
k$. Hence, the construction of $\langle G(V,E,k),\sigma(V,E,k)\rangle$
can be done in worst-case time polynomial in $\|\langle V,E,k \rangle\|$.
We next show the equivalence of the two problems.

\paragraph{Assume $\langle V,E,k \rangle \in VC$:}

Then there exists a subset $V' \subseteq V$ and a function $f: E
\rightarrow V'$ such that $|V'| \leq k$ and $\forall \langle v_m, v_n
\rangle \in E: f(\langle v_m, v_n \rangle) \in \{\langle v_m, v_n
\rangle\}$. A dependency tree for $\sigma(V,E,k)$ is constructed by:

\begin{enumerate}
\item For every $e_i \in E$,  one word $f(e_i)$ is assigned class $H_i$
and governed by $s$ in valency \meth{h$_i$}.

\item For each $v_i \in V \setminus V'$, $|E|-1$
words $v_i$ are assigned class $R$ and governed by the remaining copy of
$v_i$ in reading $U_i$ through valencies \meth{r$_1$} to
\meth{r$_{|E|-1}$}. 

\item The $v_i$ in reading $U_i$ are governed by $s$ through the
valencies \meth{u$_j$} ($j\in[1,|V|-k]$).

\item $(k-1)*|E|$ words remain in $\sigma$. These receive reading $R$
and are governed by $s$ in valencies \meth{r$_j$}
($j\in[1,\mbox{$(k-1)|E|$}]$). 
\end{enumerate}

The dependency tree rooted in $s$ covers the whole input
$\sigma(V,E,k)$. Since $G(V,E,k)$ does not give any further
restrictions this implies $\sigma(V,E,k) \in L(G(V,E,k))$ and, thus,
$\langle G(V,E,k),\sigma(V,E,k)\rangle \in DGR$.

\paragraph{Conversely assume $\langle G(V,E,k),\sigma(V,E,k)\rangle \in
DGR$:} 

Then $\sigma(V,E,k) \in L(G(V,E,k))$ holds, i.e., there exists a
dependency tree that covers the whole input. Since $s$ cannot be
governed in any valency, it follows that $s$ must be the root. The
instance $s$ of $S$ has $|E|$ valencies of class $H$, $(k-1)*|E|$
valencies of class $R$, and $|V|-k$ valencies of class $U$, whose
instances in turn have $|E|-1$ valencies of class $R$. This sums up to
$|E|*|V|$ potential dependents, which is the number of terminals in
$\sigma$ besides $s$. Thus, all valencies are actually filled. We
define a subset $V_0 \subseteq V$ by $V_0 =_{\text{def}} \{v \in V|
\exists i \in [1,|V|-k]: s.\meth{mod}(\meth{u$_i$}) =
v\}$. \mala{klk}I.e.,
\begin{equation}
\label{klk}
	|V_0| = |V| - k
\end{equation}

The dependents of $s$ in valencies \meth{h$_i$} are from the 
set $V \setminus V_0$. 
We define a  function $f: E \rightarrow V \setminus V_0$ by
$f(e_i) =_{\text{def}}
s.\meth{mod}(\meth{h$_i$})$ for all $e_i \in E$. 
By construction $f(e_i)$ is an
end point of edge $e_i$, i.e.\mala{kle}
\begin{equation}
\label{kle}
	\forall \langle v_m, v_n \rangle \in E : f(\langle v_m, v_n \rangle) \in \{ v_m, v_n \}
\end{equation}  

We define a subset $V' \subseteq V$ by $V' =_{\text{def}} \{f(e) | e \in
E\}$. Thus\mala{klf}
\begin{equation}
\label{klf}
	\forall e \in E : f(e) \in V'
\end{equation}

By construction of $V'$ and by \eqref{klk} it follows
\mala{klv}
\begin{equation}
\label{klv}
	|V'| \leq |V| - |V_0| = k
\end{equation}

From \eqref{kle}, \eqref{klf}, and \eqref{klv} we induce $\langle V,E,k
\rangle \in VC$.  \proofed

\begin{theorem}
\label{th:npc}\mala{th:npc}
The DG recognition problem is in the complexity class $\mathcal{NPC}$.
\said
\end{theorem}

The $\mathcal{NP}$-completeness of the DG recognition problem follows
directly from lemmata~\ref{th:np} and \ref{th:constr}. \proofed


%
%

\def\thisversion{$ $Revision: 1.12 $ $}
\def\thisdate{$ $Date: 1997-03-24 17:12:13+01 $ $}


\section{Conclusion
	\label{ch:conclu}}

We have shown that current DG theorizing exhibits a feature not
contained in previous formal studies of DG, namely the independent
specification of dominance and precedence constraints. This feature
leads to a $\mathcal{NP}$-complete recognition problem. The
necessity of this extension approved by most current DGs relates to
the fact that DG must directly characterize dependencies which in PSG
are captured by a projective structure and additional processes such
as coindexing or structure sharing (most easily seen in treatments of
so-called unbounded dependencies). The dissociation of tree structure
and linear order, as we have done in Section \ref{ch:grammar},
nevertheless seems to be a promising approach for PSG as well; see a
very similar proposal for HPSG \cite{Reape1989}.

The $\mathcal{NP}$-completeness result also holds for the
discontinuous DG presented in Section~\ref{ch:grammar}. This DG can
characterize at least some context-sensitive languages such as
$a^nb^nc^n$, i.e., the increase in complexity corresponds to an
increase of generative capacity. We conjecture that, provided a proper
formalization of the other DG versions presented in
Section~\ref{ch:DG}, their $\mathcal{NP}$-completeness can be
similarly shown.  With respect to parser design, this result implies
that the well known polynomial time complexity of chart- or
tabular-based parsing techniques cannot be achieved for these DG
formalisms in general.  This is the reason why the \textsc{ParseTalk}
text understanding system \cite{neuhaus+hahn96b} utilizes special
heuristics in a heterogeneous chart- and backtracking-based parsing
approach.




\end{document}